\documentclass[sigconf]{acmart}
\usepackage{pifont}
\usepackage[linesnumbered,ruled]{algorithm2e}
\usepackage{pifont}
\usepackage{multirow}
\usepackage{subcaption}
\newcommand{\cmark}{\ding{51}}
\newcommand{\xmark}{\ding{55}}
\AtBeginDocument{%
  }

\usepackage{algorithmic}
\usepackage{xcolor}
\usepackage{booktabs}
\copyrightyear{2026}
\acmYear{2026}

\setcopyright{cc}
\setcctype{by-nc-nd}
\acmConference[ICMR '26]{International Conference on Multimedia Retrieval}{June 16--19, 2026}{Amsterdam, Netherlands}
\acmBooktitle{International Conference on Multimedia Retrieval (ICMR '26), June 16--19, 2026, Amsterdam, Netherlands}
\acmDOI{10.1145/3805622.3810862}
\acmISBN{979-8-4007-2617-0/2026/06}

\begin{document}

\title[Chameleon]{Chameleon: Benchmarking Detection and Backtracking on Commercial-Grade AI-Generated Videos}


\author{Xingming Liao}
\email{liaoxingming@mails.gdut.edu.cn}
\orcid{0009-0000-9118-5968}
\affiliation{%
  \institution{School of Computer Science and Technology, Guangdong University of Technology}
  \city{Guangzhou}
  \country{China}
}

\author{Meiyu Zeng}
\email{517311281@qq.com}
\affiliation{%
  \institution{School of Computer Science and Technology, Guangdong University of Technology}
  \city{Guangzhou}
  \country{China}
}

\author{Canyu Chen}
\email{2112405115@mail2.gdut.edu.cn}
\affiliation{%
  \institution{School of Computer Science and Technology, Guangdong University of Technology}
  \city{Guangzhou}
  \country{China}
}

\author{Nankai Lin}
\authornote{Corresponding Author}
\email{neakail@outlook.com}
\affiliation{%
  \institution{School of Information Science and Technology, Guangdong University of Foreign Studies}
  \city{Guangzhou}
  \country{China}
}
% \affiliation{%
%   \institution{Guangdong Engineering Research Center of Data Security Governance and Privacy Computing}
%   \city{Guangzhou}
%   \country{China}
% }

\author{Zhuowei Wang}
\email{zwwang@gdut.edu.cn}
\affiliation{%
  \institution{School of Computer Science and Technology, Guangdong University of Technology}
  \city{Guangzhou}
  \country{China}
}

\author{Aimin Yang}
\email{amyang18@163.com}
\affiliation{%
  \institution{School of Computer Science and Technology, Guangdong University of Technology}
  \city{Guangzhou}
  \country{China}
}

% \author{Xingming Liao$^\dagger$, Meiyu Zeng$^\dagger$, Canyu Chen$^\dagger$, Nankai Lin$^{\diamondsuit*}$, Zhuowei Wang$^{\dagger*}$, Aimin Yang}

%  \author{Xingming Liao}
% \authornotemark[$^\dagger$]

% \affiliation{%
%  \institution{$^\dagger$Guangdong University of Technology, Guangzhou, China}
% }

% \affiliation{%
%  \institution{$^\diamondsuit$Guangdong University of Foreign Studies, Guangzhou, China}
% }

% \thanks{* Nankai Lin and Zhuowei Wang are co-corresponding authors.} 

\renewcommand{\shortauthors}{Xingming Liao et al.}

\begin{abstract}
The proliferation of AI-Generated Content (AIGC), especially deepfake videos, poses a severe threat to social trust by enabling fraud, privacy violations and disinformation. Existing AI-generated video detection (AGVD) benchmarks focus on open-source model generated videos, yet commercial closed-source models produce more realistic, temporally coherent videos that are underexplored in detection research. To fill this gap, we present Chameleon, a commercial-grade dataset with 1,700 AI-generated videos from 600 real-world sources across three key domains (News, Speech, Recommendation), featuring high resolution, rich annotations and 3D consistency metrics for dynamic scene spatial coherence, shifting detection from face-centric forgery to holistic scene forensics. This benchmark assesses models on two core tasks: accurate AI video detection in real-world conditions and forensic backtracking of original sources. Experimental results reveal critical limitations of existing methods in detecting and backtracking high-fidelity, spatiotemporally consistent videos from commercial closed-source models, highlighting current methods' flawed forensic reasoning and establishing Chameleon as a vital challenge for AIGC security research. The code and data are available at https://github.com/lxixim/Chameleon.
\end{abstract}





\maketitle

\begin{figure}[ht]
  \centering
   \includegraphics[width=1\linewidth]{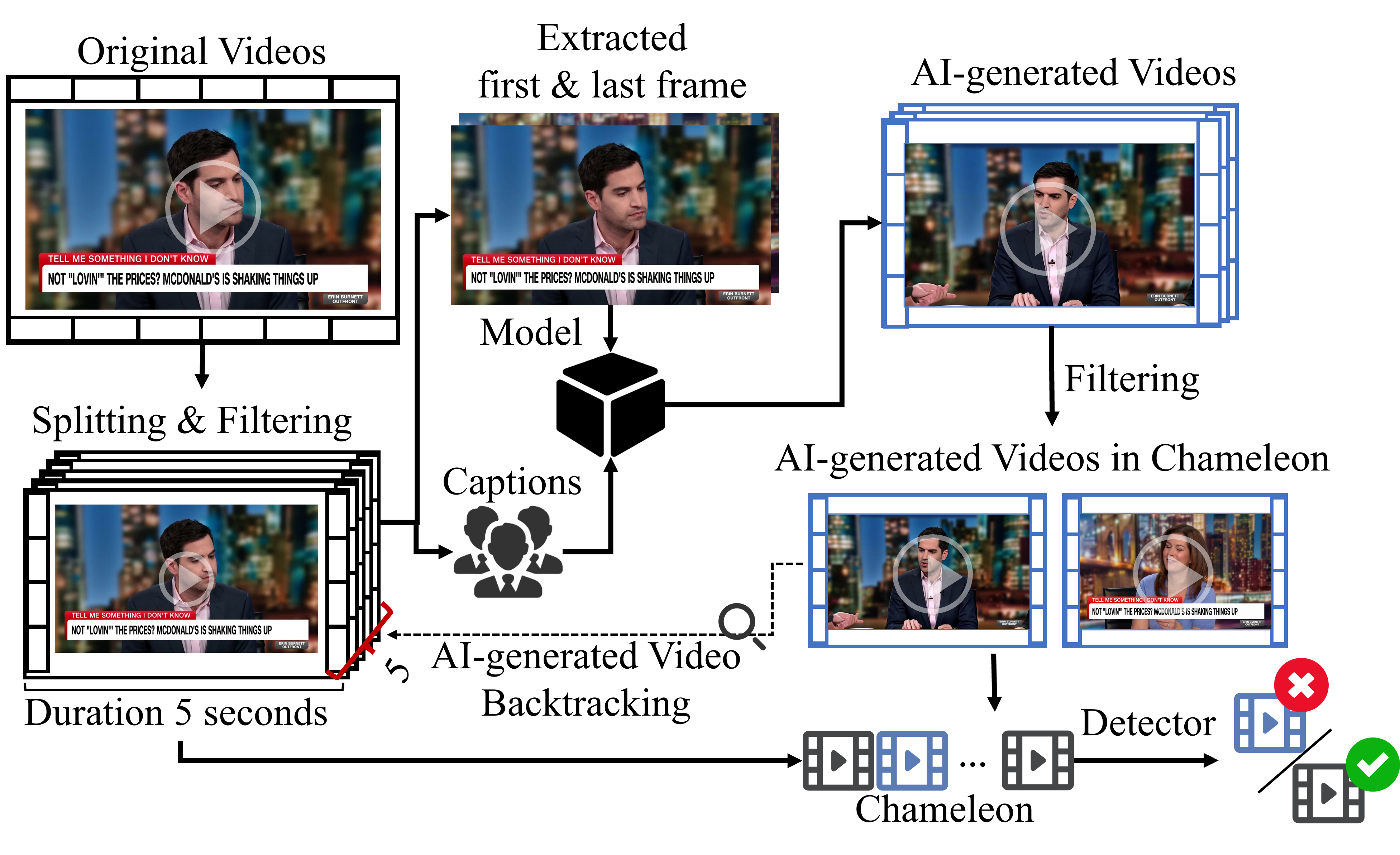}
   \caption{Introduction for the detection and backtracking of text-image to video. Blue for AI-generated video and black for real-world videos.}
   \label{fig:framework}
\end{figure}

\begin{figure*}[ht]
  \centering
   \includegraphics[width=0.9\linewidth]{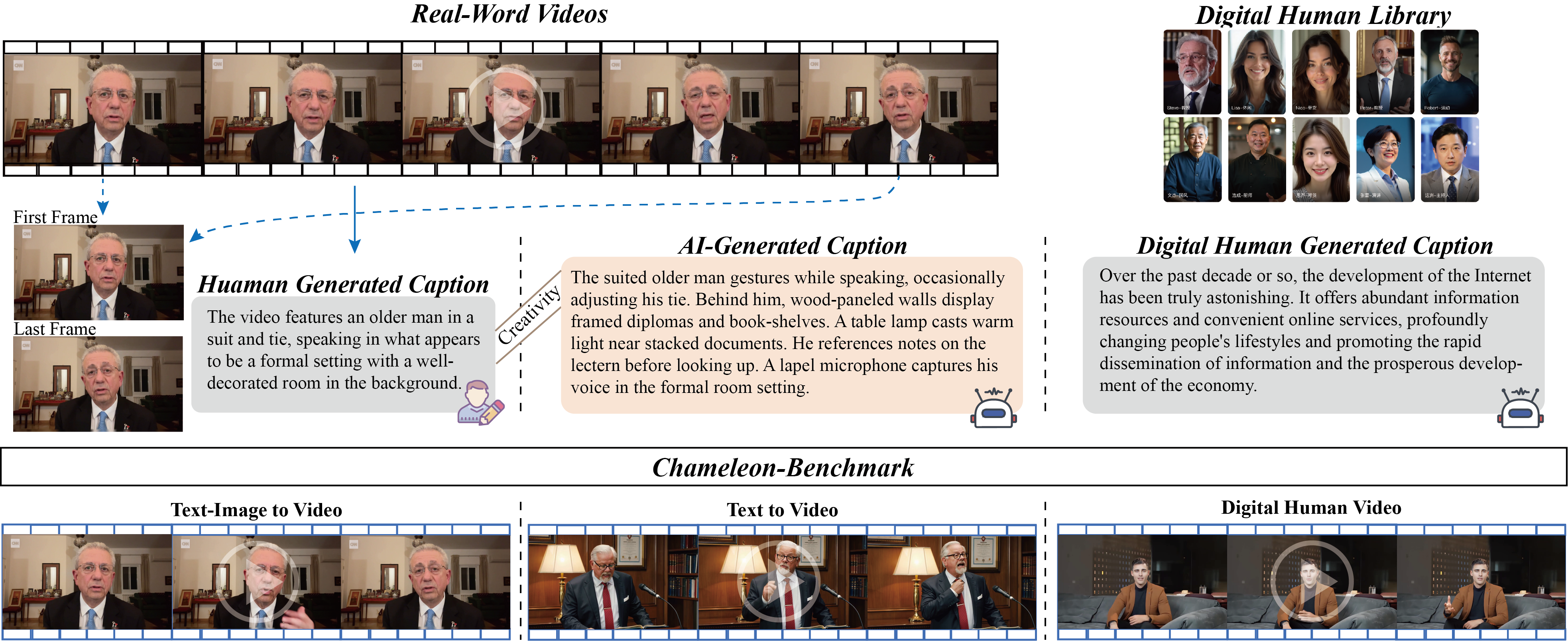}

   \caption{The construction process of text-image to video, text to video, and digital human video. The text and images for generating text-image to video are derived from real-world videos, achieving the preservation of character images and the alteration of events. The text of the text to video is derived from the video caption optimized by LLM, achieving the tampering of characters and scenes. The digital human video is derived from the existing digital human database and fabricates events.}
   \label{fig:Framework}
\end{figure*}

\section{Introduction}
While the Artificial Intelligence Generated Content (AIGC) \cite{10.1145/3576915.3623189,DBLP:journals/wc/DuLNKXSK24,Zhang2025DevelopmentCA,hua2024limitations} area has carried out a substantial amount of research on the detection of AI-generated text \cite{DBLP:journals/corr/abs-2308-05341,DBLP:conf/nips/GuoHZZFHM24,DBLP:conf/nips/Guo0J000S024} and AI-generated image \cite{DBLP:conf/nips/ZhengL0WGL024,DBLP:conf/nips/PalKPBYCH24,DBLP:conf/ccs/HaPBSSZZ24}, significant challenges persist in AI-generated video \cite{huang2024magicfight,gao2024matten}. The rapid advancement of video generation models has dramatically lowered the barrier to creating highly realistic synthetic videos, enabling malicious actors to easily forge content in high-risk scenarios such as news broadcasts, public speeches, and product recommendations. When weaponized, such technology poses critical threats to social trust, facilitating identity fraud, privacy violations, and large-scale disinformation.

In the field of AI-generated video detection (AGVD), existing works enhance the detector generalization through data augmentation \cite{wang2021representative,wang2020cnn}, GAN-based architectures \cite{tan2024frequency}, reconstruction techniques \cite{cao2022end}, and adversarial training \cite{chen2022self}, achieving promising 
performance on videos involving facial replacement and single-object scenarios. To evaluate these methods, various benchmarks have been proposed. Deepfake benchmarks such as Faceforensics++ \cite{DBLP:conf/iccv/RosslerCVRTN19}, DFDC \cite{DBLP:journals/corr/abs-1910-08854}, Wilddeepfake \cite{zi2020wilddeepfake}, and CoReD \cite{kim2021cored} employ facial forgery and tampering operations based on CNNs or GANs, providing relatively rich facial manipulation features. To address the limitation of single and fixed facial scenes, CDDB \cite{Li_2023_WACV} extends the forgery scope from facial regions to more general content. More recently, newly developed benchmarks such as GenVidBench \cite{ni2025genvidbench} and GenWorld \cite{chen2025genworld} incorporate videos from generative models (e.g., SVD, CogVideo), expanding coverage to broader visual content.

However, commercial platforms (e.g., Runway, Kling, Sora \cite{DBLP:journals/corr/abs-2402-17177}) employ closed-source models trained on larger-scale proprietary datasets with 
substantially greater computational resources, producing videos with superior realism, temporal coherence, and complex scene transitions that remain inadequately covered by existing benchmarks. Furthermore, while current methods focus on binary authenticity classification, the critical task of tracing AI-generated videos back to their source materials, which is essential for investigating identity theft and privacy violations, remains unexplored. These observations raise crucial questions: \textit{1) Can existing detection methods accurately identify videos generated by commercial closed-source models? 2) Which real video is used to generate this particular AI-generated video?}

To fill these gaps, we construct an AGVD benchmark \textbf{Chameleon} using multiple 
commercial video generation platforms and systematically evaluate a series of detection methods. Chameleon uniquely provides native high-resolution outputs and 3D consistency, producing videos that maintain coherent spatial geometry and physically reasonable shadows in dynamic scenes. This shifts detection research beyond face-centric forgery toward holistic content verification. Specifically, as shown in Figure \ref{fig:Framework}, for known data sources, we collect real videos from three domains that are vulnerable to tampering and prone to wide dissemination: news broadcasts, public speeches, and product recommendations. Based on captions and keyframes extracted from these real videos, we construct AI-generated videos through two pipelines: text-to-video and text-image-to-video. Since these generated videos maintain direct correspondence with their real sources, we introduce a backtracking task to trace AI-generated videos to their original materials, enabling forensic investigation of identity theft and content misappropriation. For unknown data sources, we further generate digital human videos using a digital human library. As shown in Table \ref{tab:dataset_comparison}, Chameleon inherits core features of existing benchmarks, including scene diversity, video pairs, generation prompts, semantic labels, and domain variety. In summary, this paper makes three main contributions:

\begin{itemize}
    \item We construct Chameleon, a new AGVD benchmark featuring videos generated 
    by commercial closed-source models, presenting more realistic and challenging 
    samples for detection research.
    \item We explore the performance of different detection methods and backtracking backbone in Chameleon, evaluating and understanding their capabilities and limitations.
    \item This work bridges the gap between AGVD benchmark construction and real-world forensic needs, establishing new challenges to combat evolving 
    AI-generated video threats.
\end{itemize}

\begin{table*}[ht]
\small
\caption{Comparison with existing AGVD benchmarks. Chameleon surpasses other benchmarks by incorporating native high-resolution outputs and 3D consistency from commercial video generators. This enables comprehensive evaluation of detection methods against videos with superior realism and spatial coherence that closely mimic real-world content.}
\begin{tabular}{lccccccc}
\hline
\textbf{Dataset} & \textbf{Scene Diversity} & \textbf{Video Pairs} & \textbf{Prompt/Image} & \textbf{Semantic Label} & \textbf{Domain Variety} & \textbf{3D Consistency} & \textbf{High-Res Native} \\ \hline
FF++             & \xmark                        & \cmark                    & \xmark                     & \xmark                       & \xmark                       & -                       & \xmark                        \\
DFDC             & \cmark                        & \cmark                    & \xmark                     & \xmark                       & \xmark                       & -                       & \xmark                        \\
Wilddeepfake     & \cmark                        & \xmark                    & \xmark                     & \xmark                       & \xmark                       & -                       & \xmark                        \\
CoReD            & \xmark                        & \cmark                    & \xmark                     & \xmark                       & \xmark                       & -                       & \xmark                        \\
CDDB             & \cmark                        & \xmark                    & \xmark                     & \xmark                       & \xmark                       & \xmark                       & \xmark                        \\
GVD              & \cmark                        & \xmark                    & \xmark                     & \xmark                       & \xmark                       & \xmark                       & \xmark                        \\
GVF              & \cmark                        & \cmark                    & \cmark                     & \cmark                       & \xmark                       & \xmark                       & \xmark                        \\
GenVideo         & \cmark                        & \xmark                    & \xmark                     & \xmark                       & \xmark                       & \xmark                       & \xmark                        \\
GenVidBench      & \cmark                        & \cmark                    & \cmark                     & \cmark                       & \cmark                       & \xmark                       & \xmark                        \\ \hline
\textbf{Chameleon}        & \cmark                       & \cmark                    & \cmark                     & \cmark                       & \cmark                       & \cmark                       & \cmark                        \\ \hline
\end{tabular}
\label{tab:dataset_comparison}
\end{table*}

\section{Related Works}
\subsection{Deepfake Detection Benchmarks}
Recent research on AGVD datasets has focused primarily on three categories. The first category includes face-centric datasets such as DFDC \cite{DBLP:journals/corr/abs-1910-08854}, Celeb-DF \cite{DBLP:conf/cvpr/LiYSQL20}, 
and Faceforensics++ \cite{DBLP:conf/iccv/RosslerCVRTN19}. These focus on face replacement and expression manipulation, generating fake content through synchronous 
changes in faces or mouth regions within fixed camera views. Although these datasets 
are large-scale and existing methods achieve promising performance, they are limited 
by their single-scene nature and reliance on local facial features 
\cite{Wang_2020_CVPR}. The second category comprises early generalized AI-generated datasets such as ForenSynths and VideoGPT \cite{DBLP:journals/corr/abs-2104-10157}. These expand tampering types to include gestures and background replacements while introducing temporal dynamics \cite{guillaro2023trufor}. However, they still struggle with accurately simulating real attack patterns and lack sufficient domain coverage, often focusing solely on single objects or faces. The third category emerges with recent open-source video generation models. GenVidBench \cite{ni2025genvidbench} constructs a large-scale benchmark using open-source generators such as SVD and CogVideo, providing semantic labels and generation prompts for comprehensive analysis. GenWorld \cite{chen2025genworld} further incorporates cross-prompt diversity, including text-to-video, image-to-video, and video-to-video generation pipelines. While these benchmarks advance detection research beyond face-centric scenarios, they predominantly rely on open-source models whose outputs often exhibit detectable artifacts.

However, commercial platforms employ closed-source models trained on larger-scale proprietary datasets with greater computational resources, producing videos with superior realism and temporal coherence that remain inadequately covered. To fill this gap, we construct Chameleon, a commercial-grade AGVD benchmark that features native high-resolution outputs and 3D consistency, providing a more challenging testbed for detection research.

\subsection{Deepfake Detection Methods}

AGVD mainly focuses on two directions: data-driven feature analysis \cite{li2018ictu,ciftci2020fakecatcher} and deep-learning based methods \cite{she2024using,kamat2023revisiting}. Nowadays, detection techniques mainly focus on deep-learning based methods. Early deep-learning based methods are mostly based on convolutional neural networks to extract local texture features. Tan et al. \cite{DBLP:conf/cvpr/TanLZWGLW24} proposed that artifacts introduced by the up-sampling operation can be reversed and applied to detection. In addition, they also proposed a frequency domain sensing detection framework \cite{tan2024frequency}. To meet the real-time detection requirements, Romeo et al. \cite{lanzino2024faster} designed a lightweight detector based on a binary neural network, providing a feasible solution for end-to-end deployment. 

With the diversification of generation technologies, researchers are beginning to explore a broader AI-generated detection paradigm. Davide et al. \cite{DBLP:conf/cvpr/CozzolinoPCNV22} proposed a lightweight detection strategy based on CLIP features \cite{zhang2022pointclip}. Gu et al. \cite{DBLP:conf/aaai/GuCYDLM22} designed a dynamic inconsistency Learning model to capture spatiotemporal anomalies from densely sampled video footage. With the emergence of LVMs, they have performed well in cross-modal understanding \cite{10.1145/3672758.3672824,DBLP:journals/corr/abs-2304-03086}. Jia et al. \cite{DBLP:journals/corr/abs-2403-14077} proposed to use LVMs \cite{zhang2024gpt,team2023gemini} to detect forged face images, and designed text prompts.

However, the above methods still face two major challenges: First, the existing datasets are not diverse in scene and domain diversity, resulting in the model overfitting specific artifacts; Second, the generalization ability of the detection algorithm to new generation models (such as Runway and Kling) \cite{zhang2023toward,huang2025m4v} is insufficient. Muhammad et al. \cite{DBLP:journals/corr/abs-2401-02418} further explored the performance of advanced LVMs in the AGVD task by introducing prompt learning, revealing the potential of LVMs in the AGVD task and providing a new idea for building a robust inspection system.

\section{Chameleon}

In this section, we introduce our self-constructed AGVD benchmark, Chameleon, which considers the AI-generated video's scene diversity and domain variety. Subsection 3.1 details the selection of source data used for the dataset. Subsection 3.2 describes the strategy employed to construct AI-generated videos. Finally, Subsection 3.3 outlines the dataset’s overall composition and splitting protocol, emphasizing balanced thematic coverage and a standardized train/validation/test partition to enable fair, systematic evaluation.

\subsection{Source Data and Preprocessing}

In this work, we select real videos from news broadcasts (News), public speeches (Speech), and product recommendations (Recommendation) as raw videos for text-image to video and text to video, as these domains represent high-risk scenarios for misinformation, identity impersonation, and commercial fraud, thereby enabling a focused study on safeguarding public trust and personal privacy against realistic threats. Specifically, the source data are selected from three public video platforms, namely CNN, TED, and YouTube. The raw data covered 120 videos, each of News, Speech, and Recommendation videos. Additionally, these videos have a duration ranging from 2 to 20 minutes.

Raw videos are first segmented into non-overlapping 5-second video segments using FFmpeg. To ensure the selected videos exhibited scene diversity and domain variety, we manually fliter the segments based on two criteria: \textbf{(1) the presence of scene transitions} such as abrupt cuts in news broadcasts, slide changes in speeches, or environmental shifts in product recommendations and \textbf{(2) diversity in domains} like interactions between humans, objects, and backgrounds. Three annotators independently score each video, retaining only the top 5 video segments per raw video. After aggregation, the final dataset comprises 200 videos per category (600 total), denoted as $\mathcal{V}^R$.

\subsection{AI-generated Video Creation}

Current LVMs, including Runway Gen 3\footnote{https://runwayml.com/}, Jimeng\footnote{https://jimeng.jianying.com/ai-tool/home/}, and Kling\footnote{https://klingai.kuaishou.com/} (Due to strict restrictions on character generation, Sora is excluded), simultaneously generate AI-generated videos based on the provided text and images, and eventually obtain 600 AI-generated videos in which the identity of the character is maintained but the scene action is tampered with. Constrained by the original scene shown in the real first and last frames, the generated content needs to be tampered with within the ``real framework''. The AI-generated videos have a direct connection with real videos and are used for video backtracking. The construct process for text-image to video and the backtracking task are shown in Figure \ref{fig:framework}.

\textbf{Text-Image to Video.} We take the following steps to obtain the text and pictures needed to generate the AI-generated videos. To generate text for each video in $\mathcal{V}^R$, we invite three professional researchers with multi-modal knowledge to provide a detailed textual description of the video from three core aspects: Key Objects, Actions, and Scene Transitions. Two experts carry out the annotation, and when there is ambiguity about the description of the video, it will be annotated by a third expert. Next, to get the images for creation, the first and last frames of $\mathcal{V}^R$ are extracted using FFmpeg to serve as anchor points for temporal consistency. Finally, the caption and first and last frames are regarded as metadata to guide the LVMs to generate the video.

\textbf{Text to Video.} To further enhance the capability of LVMs in creatively generating AI-generated videos to simulate potential manipulations of character identities and overall scenes in the real world, we rely on text descriptions of videos for generation. Additionally, to enrich these video text descriptions, we utilize LLMs to process the original texts. Ultimately, 600 text descriptions are selected to generate 600 AI-generated videos.

\textbf{Digital Human Video.} The generation of digital human videos takes ``digital human library anchoring and instruction-driven generation'' as its core. First, scenes and public digital humans are selected from the digital human library. Then, LVMs generates instructions containing action voices. Finally, the dynamic and background scenes of digital humans are generated and fused for rendering, obtaining 500 digital human videos.

\subsection{Data Contents}
 
In total, the constructed dataset comprises 2,300 videos, all standardized to a frame rate of 30 FPS.  Detailed data are shown in Table \ref{tab:dataset_distribution}. To systematically evaluate the performance of the model, we distribute the three themes and ultimately divide the dataset, validation set, and test set in a 6:2:2 ratio.

\section{Task Definition}
\subsection{Formal Problem Description}
In the task of detecting AI-generated video, the objective is to develop a binary classification detector capable of accurately identifying AI-generated video. We consider a collection of real-world and AI-generated video sampled from various sources, denoted as $\mathcal{V}^{R} = \{\mathcal{V}_1^{R}, \mathcal{V}_2^{R}, ..., \mathcal{V}_n^{R}\}$ and $\mathcal{V}^{F} = \{\mathcal{V}_1^{F}, \mathcal{V}_2^{F}, ..., \mathcal{V}_n^{F}\}$, respectively, where the $n$ represents the number of the videos. The purpose of the detector is to output the corresponding label  $\mathcal{Y}^R=\{\mathcal{Y}_1^R, \mathcal{Y}_2^R,..., \mathcal{Y}_n^R\}$ and $\mathcal{Y}^F=\{\mathcal{Y}_1^F, \mathcal{Y}_2^F,..., \mathcal{Y}_n^F\}$ for each video, where $\mathcal{Y}=1$ indicates the AI-generated video and $\mathcal{Y}=0$ indicates the real-world video. In this study, we also propose a backward tracking task for AI-generated video aimed at identifying the real-world video used to generate that AI-generated video. Specifically, the labels of real-world videos are denoted as $\mathcal{L}^R=\{\mathcal{L}_1^R,\mathcal{L}_2^R,...,\mathcal{L}_n^R\}$, for AI-generated video, their labels are denoted as  $\mathcal{L}^F=\{\mathcal{L}_1^F,\mathcal{L}_2^F,...,\mathcal{L}_n^F\}$. As for an AI-generated $\mathcal{V}_i^{F}$ $(i
\in{1,2,...,n})$, the real-world video from which it comes is represented as $\mathcal{L}_{i}^F\in{\{\mathcal{L}_1^R,\mathcal{L}_2^R,...,\mathcal{L}_n^R\}}$.

\begin{table}[t]
\centering
\caption{Detailed distribution of the number of videos. It includes three themes: News(N), Speech(S), and Recommendation(R).}

\begin{tabular}{l|lccc}
\hline
                       & \textbf{Type}& \textbf{Theme}& \textbf{Quantity}& \textbf{Total}\\ \hline
\multirow{3}{*}{Real}  & \multirow{3}{*}{Video Segments}      & N            & 200      & \multirow{3}{*}{600} \\
                       &                                      & S          & 200      &                      \\
                       &                                      & R & 200      &                      \\ \hline
\multirow{7}{*}{Fake}  & \multirow{3}{*}{Text-Image to Video} & N            & 200      & \multirow{3}{*}{600} \\
                       &                                      & S          & 200      &                      \\
                       &                                      & R & 200      &                      \\
                      \cline{2-5}& \multirow{3}{*}{Text to Video}       & N            & 200      & \multirow{3}{*}{600} \\
                       &                                      & S          & 200      &                      \\
                       &                                      & R & 200      &                      \\
                      \cline{2-5}& Digital Human Video                  & -               & 500      & 500                  \\ \hline
\end{tabular}

\label{tab:dataset_distribution}
\end{table}

\subsection{AI-generated Detection}
For an input video $\mathcal{V}_i$, the detector first extracts the video into frames. Extract a sequence of frames from $\mathcal{V}_i$ at a fixed sampling rate $r$ ($r=2$ frames per second):
\begin{equation}{\mathcal{S}_i}=\left\{\mathcal{I}_{i_t}\mid t\in\{1,2,\ldots,k\}\right\},\quad k=\lfloor T\cdot r\rfloor,
\end{equation}
where $\mathcal{I}_{t}\in\mathbb{R}^{H\times W\times3}$ represents the $t$-th frame with spatial resolution $H\times W$. $T$ represents the duration of the video. A video contains $k$ frames, converting the continuous video stream into a discrete frame sequence for computational processing. For each extracted frame $\mathcal{I}_{i_t} \in {\mathcal{S}_i}$, a binary classifier $f_\theta:\mathbb{R}^{H\times W\times3}\to\{0,1\}$ predicts its authenticity:
\begin{equation}
\small
\mathcal{L}_{\text{frame}} = -\frac{1}{K} \sum_{t=1}^{K} \left[ w \cdot y_{i_t} \log(p_{i_t}) + w' \cdot (1 - y_{i_t}) \log(1 - p_{i_t}) \right].
\end{equation}
\begin{equation}
\hat{y}_{i_t}= f_\theta(\mathcal{I}_{i_t}), \quad \text{where}
\begin{cases}
\hat{y}_{i_t} = 1, &  p_{i_t} \geq \tau, \\
\hat{y}_{i_t} = 0, &  p_{i_t} < \tau,
\end{cases}
\end{equation}
where $ p_{i_t}$ represents the likelihood of frame being AI-generated and $\tau$ represents the threshold. $w$ and $w'$ represent the learnable weights. $y_{i_t}$ represents the true label of the $t$-th frame. Specifically, each video frame $\hat{y}_{i_t}$ is labeled with 0 or 1, indicating whether it belongs to the category of real-world video ($\mathcal{Y}=0$) or AI-generated video ($\mathcal{Y}=1$). 

To aggregate frame-level predictions into a video-level authenticity decision, we count the number of correctly predicted frames in the video and calculate their proportion, and then convert it into a video-level decision:
\begin{equation}
\mathcal{P}_{i} = \frac{\sum_{t=1}^k \mathbb{I}(y_{i_t}= \hat{y}_{i_t})}{k},
\end{equation}
where $\mathcal{P}_{i}$ denotes the correct prediction rate for all frames in the $i$-th video, namely, confidence probability.  $\mathbb{I}$(·) is an indicator function that returns 1 if the condition in parentheses holds, and 0 otherwise. Further, for the video detector, the calculation is as follows:
\begin{equation}
\hat{\mathcal{Y}_i}^{F} = 
\begin{cases}
1, &  \mathcal{P}_{i} \geq \theta, \\
0, & \mathcal{P}_{i} < \theta,
\end{cases}
\end{equation}
\begin{equation}
\hat{\mathcal{Y}_i}^{R} = 
\begin{cases}
0, &  \mathcal{P}_{i} \geq \theta, \\
1, & \mathcal{P}_{i} < \theta.
\end{cases}
\end{equation}

Specifically,  when the prediction probability $\mathcal{P}_{i}$ of the whole video exceeds the confidence threshold $\theta$, AI-generated video will be labeled as 1, and real-world videos will be labeled as 0; otherwise, AI-generated and real-world videos will be incorrectly labeled as 0 and 1, respectively.

\subsection{AI-generated Video Backtracking}

The AI-generated video backtracking task aims to identify the unique real-world source footage from which an AI-generated video originates. Formally, given an AI-generated video query, the objective is to retrieve its corresponding authentic source from a gallery of real-world videos. Unlike standard video retrieval, this task focuses on provenance verification, requiring the model to establish a definitive one-to-one correspondence between the synthesized content and its ground-truth source.

\textbf{Simulating Real-world Evasion via Temporal Pruning.}
In real-world scenarios, AI-generated videos circulated on social media often undergo post-processing to evade detection or conform to platform constraints. Common anti-forensic techniques include trimming the start or end of a video, which disrupts temporal alignment and compromises watermark-based or frame-by-frame matching methods. 
To simulate adversarial conditions and evaluate the robustness of backtracking models against anti-forensic techniques (e.g., removing watermarks or intro/outro sequences), we introduce a Temporal Pruning Protocol. Specifically, for the query videos in the test set, we apply random temporal cropping operations: removing the first $N$ frames, the last $N$ frames, or both. This setup creates a partial overlap scenario between the query and the gallery, forcing the model to rely on robust spatiotemporal features rather than exact frame-to-frame alignment.

\textbf{Video Feature Extraction.}
Let $\mathcal{V}_i^{F}$ denote an AI-generated query video, and let $\{\mathcal{V}_j^{R}\}_{j=1}^{n}$ denote a gallery of real-world videos. A video encoder $\text{VideoEncoder}(\cdot)$ is employed to map each video into a compact representation in a shared embedding space:
\begin{equation}
\mathbf{v}_i^{F} = \text{VideoEncoder}(\mathcal{V}_i^{F}),
\end{equation}
\begin{equation}
\mathbf{v}_j^{R} = \text{VideoEncoder}(\mathcal{V}_j^{R}), \quad j \in \{1, 2, \ldots, n\}.
\end{equation}
where $\mathbf{v}_i^{F} \in \mathbb{R}^{d}$ and $\mathbf{v}_j^{R} \in \mathbb{R}^{d}$ denote video-level embeddings for AI-generated video and real-world videos, respectively. The video encoder aggregates information throughout the duration of the video, enabling the representation to capture both semantic consistency and temporal dynamics, which are essential for reliable source identification.

\textbf{Similarity-based Backtracking.}
Given the embedding of a query video and those of the gallery videos, backtracking is performed by measuring similarity in the embedding space. The similarity between $\mathcal{V}_i^{F}$ and $\mathcal{V}_j^{R}$ is computed using cosine similarity:
\begin{equation}
\text{Sim}(\mathcal{V}_i^{F}, \mathcal{V}_j^{R}) 
= 
\frac{\mathbf{v}_i^{F} \cdot \mathbf{v}_j^{R}}
{\|\mathbf{v}_i^{F}\| \, \|\mathbf{v}_j^{R}\|}.
\end{equation}

All real-world videos in the gallery are ranked according to their similarity to the query video. The backtracking result is defined as the real-world video with the highest similarity score:
\begin{equation}
j^* = \arg\max_{j \in \{1,\ldots,n\}} \text{Sim}(\mathcal{V}_i^{F}, \mathcal{V}_j^{R}).
\end{equation}

By formulating AI-generated video backtracking as a retrieval problem in a unified video embedding space, the proposed benchmark directly targets the identification of the unique real-world source video for a given AI-generated video.

\begin{table}[]
\caption{Results of various state-of-the-art methods trained on Chameleon.}
\begin{tabular}{lccc}
\hline
Model            & Detection Level & ACC    & AUC\\ \hline
NPR              & Image           & 0.6300     & 0.8748\\
STIL             & Video           & 0.6074     & 0.7933\\
TALL             & Video           & 0.7726     & 0.8507\\
DeMamba          & Video           & 0.8069     & 0.9263\\ \hline
Gemini-1.5-flash & Image           & 0.3475     & 0.4262\\
Gemini-2.5-pro   & Image           & 0.4275     & 0.4820\\
GPT-4.1          & Image           & 0.4799     & 0.6606\\
GPT-4o           & Image           & 0.5725     & 0.7592\\
Qwen-VL-max      & Image           & 0.4750     & 0.6193\\
Claude-3.5       & Image           & 0.3367     & 0.5404\\
Claude-3.7       & Image           & 0.2368     & 0.5539\\ \hline
\end{tabular}
\label{tab:base_results}
\end{table}

\begin{table}[t]
\centering
\caption{Accuracy of different methods for detecting AI-generated and real-world videos in the Chameleon dataset under varying confidence thresholds.}
\setlength{\tabcolsep}{0.7mm}
\begin{tabular}{lccccc}
    \hline
    \multirow{2}{*}{\textbf{Method}} & \multicolumn{5}{c}{\textbf{Chameleon}}           \\ \cline{2-6}
                                 & $acc@0.6$ & $acc@0.7$ & $acc@0.8$ & $acc@0.9$ & $acc@1.0$ \\ \hline
NPR                              & 0.5975& 0.5650& 0.5500& 0.5100& 0.4325\\
Gemini-2.5-pro                   & 0.3950& 0.3400& 0.2900& 0.2300& 0.1425\\
Gemini-1.5-flash                 & 0.2875& 0.2525& 0.2150& 0.1900& 0.1350\\
GPT-4.1                           & 0.4598& 0.4397& 0.4070& 0.3719& 0.3241\\
GPT-4o                           & 0.5425& 0.4800& 0.4200& 0.3550& 0.2575\\
Qwen-VL-max                      & 0.4575& 0.4500& 0.4350& 0.4300& 0.4025\\
Claude 3.7                       & 0.2368& 0.2340& 0.2340& 0.2284& 0.2173\\
Claude 3.5                       & 0.3241& 0.3141& 0.3065& 0.2965& 0.2688\\ \hline
\end{tabular}
\label{tab:accuracy_methods}
\end{table}

\begin{table*}[]
\centering
\small
\caption{Accuracy of different models under varying confidence thresholds for backtracking task and the optimal result is \textbf{bolded}. Three themes: News(N), Speech(S), and Recommendation(R).}
\resizebox{\textwidth}{!}{
\begin{tabular}{llllllllllllll}
\hline
\multicolumn{2}{l}{\multirow{2}{*}{Backbone}}                & \multicolumn{12}{c}{Chameleon}                                                                                                                                                                            \\ \cline{3-14} 
\multicolumn{2}{l}{}                                         & $Mean_{R@1}$      & $Mean_{R@3}$      & $Mean_{MRR}$      & $N_{R@1}$         & $N_{R@3}$         & $N_{MRR}$         & $R_{R@1}$         & $R_{R@3}$         & $R_{MRR}$         & $S_{R@1}$         & $S_{R@3}$         & $S_{MRR}$         \\ \hline
\multirow{2}{*}{CLIP}   & clip-vit-b16                     & 0.683          & 0.883          & 0.795          & 0.800          & 0.950          & 0.883          & 0.475          & 0.825          & 0.670          & \textbf{0.775} & \textbf{0.875} & \textbf{0.830} \\
                        & clip-vit-b32                     & 0.650          & 0.883          & 0.773          & 0.725          & 0.950          & 0.846          & 0.500          & 0.850          & 0.681          & 0.725          & 0.850          & 0.794          \\
\multirow{2}{*}{DINO}   & dino-deit-small-patch16         & \textbf{0.733} & 0.883          & \textbf{0.821} & 0.800          & 0.950          & 0.887          & \textbf{0.675} & 0.925          & \textbf{0.809} & 0.725          & 0.775          & 0.766          \\
                        & dino-vit-base-patch16           & \textbf{0.733} & 0.892          & \textbf{0.821} & \textbf{0.825} & 0.950          & \textbf{0.900} & 0.625          & 0.925          & 0.780          & 0.750          & 0.800          & 0.783          \\
\multirow{2}{*}{DINOv2} & dinov2-vitb14                     & 0.717          & 0.892          & 0.808          & 0.775          & 0.950          & 0.871          & 0.625          & 0.900          & 0.763          & 0.750          & 0.825          & 0.791          \\
                        & dinov2-vits14                     & 0.708          & 0.875          & 0.802          & 0.800          & 0.925          & 0.880          & 0.625          & 0.925          & 0.782          & 0.700          & 0.775          & 0.745          \\
\multirow{3}{*}{VIT}    & vit-base-patch16-224        & 0.725          & \textbf{0.900} & 0.815          & \textbf{0.825} & 0.900          & 0.881          & 0.625          & \textbf{0.975} & 0.788          & 0.725          & 0.825          & 0.776          \\
                        & vit-base-patch16-224-in21k & 0.650          & 0.858          & 0.761          & 0.725          & 0.875          & 0.816          & 0.450          & 0.900          & 0.674          & 0.775          & 0.800          & 0.792          \\
                        & vit-base-patch32-224-in21k & 0.650          & 0.892          & 0.775          & 0.700          & \textbf{0.975} & 0.826          & 0.525          & 0.850          & 0.712          & 0.725          & 0.850          & 0.787          \\ \hline
\end{tabular}
}
\label{tab:backtracking_backbone}
\end{table*}

\section{Benchmarking Results}
\subsection{Evaluation Metric}

To comprehensively evaluate the proposed benchmark, multiple evaluation metrics are adopted to assess its performance from different perspectives. For AI-generated Content Detection, the evaluation focuses on the correctness and discriminative capability of the detector. For the video backtracking task, the evaluation emphasizes the accuracy and ranking quality of identifying the unique real-world source video corresponding to an AI-generated query.

\noindent\textbf{Metrics for AI-generated Content Detection.}
Following established practices in prior works~\cite{DBLP:conf/cvpr/TanLZWGLW24, lanzino2024faster}, we employ Accuracy (ACC) and the Area Under the Receiver Operating Characteristic Curve (AUC) to evaluate detection performance. ACC measures the overall correctness of binary classification, with the accuracy calculation based on a threshold value of 0.5. AUC assesses the detector's discriminative capability across all decision thresholds, providing a threshold-independent evaluation of model performance.

\noindent\textbf{Metrics for AI-generated Video Backtracking.}
For the backtracking task, the objective is to retrieve the unique real-world source video corresponding to an AI-generated query from a gallery of candidate videos. To evaluate both exact source identification and ranking quality, retrieval-based metrics are adopted. Given the limited scale of our benchmark and the nature of the one-to-one matching task, the evaluation focuses on top-ranked retrieval performance. Specifically, Recall at $K$ ($R@K$) is employed to measure the percentage of queries for which the ground-truth source video appears within the top-$K$ retrieved results, with $K$ set to 1 and 3 to assess exact match accuracy and top-3 retrieval performance, respectively. Mean Reciprocal Rank (MRR) is also reported, which computes the average of the reciprocal ranks of the ground-truth source videos across all queries, providing a comprehensive evaluation of ranking quality.

\begin{figure*}[h]  
  \centering
  \includegraphics[width=0.85\linewidth]{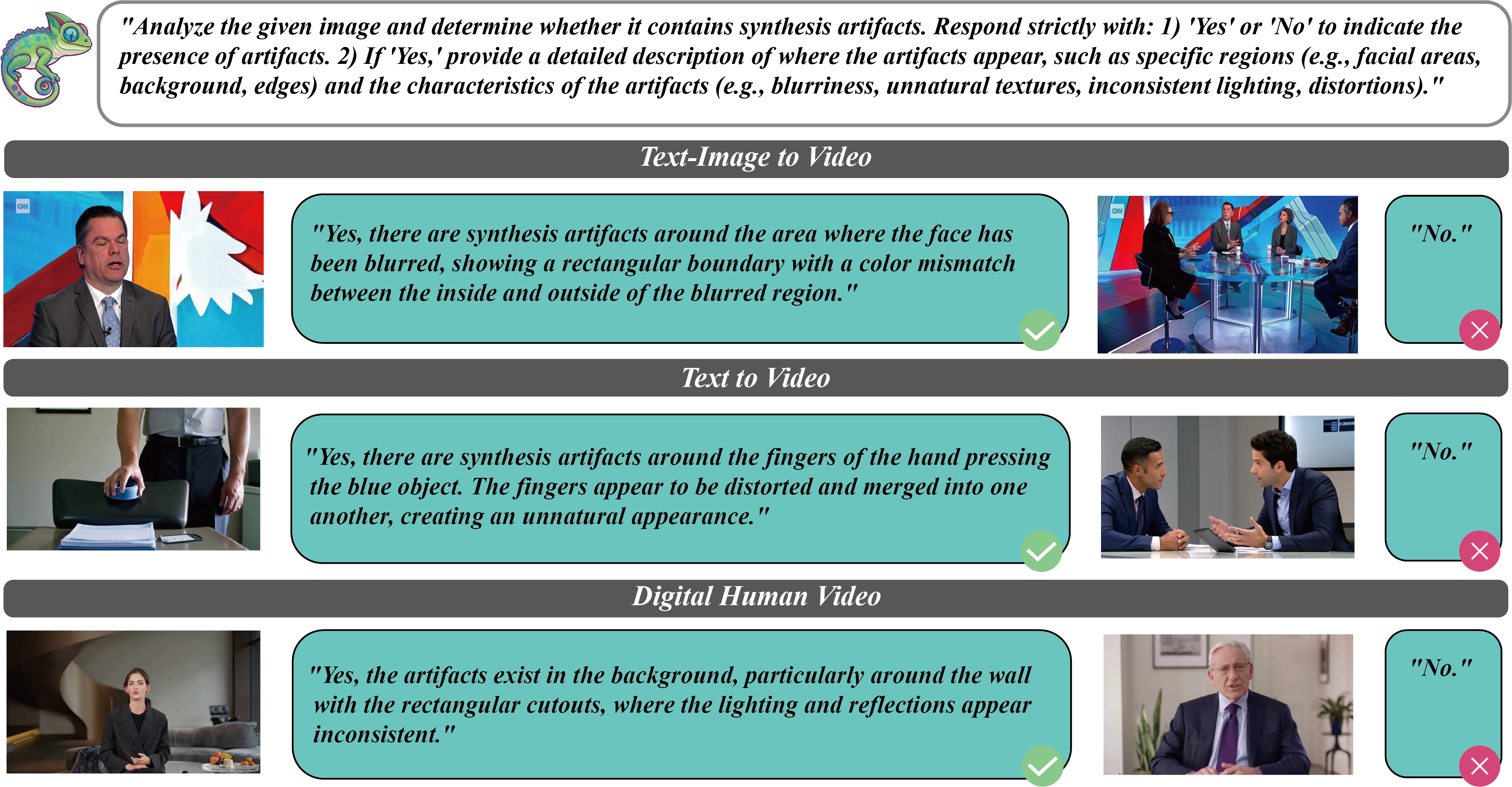}

  \caption{Examples of GPT-4o detection of three types of AI-generated videos in the Chameleon dataset. The left in the figure are AI-generated videos with correct predictions, while the ones on the right are AI-generated videos with incorrect predictions.
  }
  \label{fig:LVM}
\end{figure*}

\subsection{Baseline}
To comprehensively evaluate the proposed dataset, we conduct extensive experiments on two distinct tasks: AGVD and Backtracking.

\noindent\textbf{Video Detection Baselines.} 
We benchmark the performance of video detection using two categories of approaches: specialized deep learning-based methods and general-purpose LVMs.
For deep learning-based baselines, we compare four methods: NPR~\cite{DBLP:conf/cvpr/TanLZWGLW24}, STIL~\cite{gu2021spatiotemporal}, TALL~\cite{xu2023tall}, and DeMamba~\cite{chen2024demamba}. 
To assess the zero-shot capabilities of current foundation models for the AGVD task, we select seven advanced LVMs: Gemini-2.5-pro~\cite{team2023gemini}, Gemini-1.5-flash~\cite{team2023gemini}, GPT-4.1~\cite{zhang2024gpt}, GPT-4o~\cite{islam2024gpt}, Qwen-VL-max, Claude 3.7, and Claude 3.5~\cite{kurokawa2024diagnostic}. 
Following the methodology of Jia et al.~\cite{DBLP:journals/corr/abs-2403-14077}, we design task-specific prompts to evaluate these LVMs on samples across various scenes and domains.

\noindent\textbf{Backtracking Baselines.} 
For the backtracking task, which aims to identify the specific source generator of the AI-synthesized videos, we employ several mainstream vision backbones as feature extractors. Specifically, we utilize CLIP \cite{radford2021learning}, DINO \cite{zhang2022dino}, DINOv2 \cite{oquab2023dinov2}, and VIT \cite{dosovitskiy2020image} to extract discriminative features from video frames. These visual representations are then used to classify the source model of the generated content.

\subsection{Implementation Details}
All the experimental settings of the baseline for deep learning methods are kept consistent with the reference. Specifically, all models are trained with a consistent batch size of 32 using the Adam optimizer. For the AI-generated video backtracking task, we strictly follow the Temporal Pruning Protocol defined in the problem formulation. In our experiments, we set the pruning parameter $N=16$. This means that for every query video in the test set, we randomly remove the initial 16 frames, the final 16 frames, or both, simulating realistic video processing artifacts. The confidence threshold $\tau$ in Equation 3 is empirically set to 0.5, following standard evaluation protocols in video detection. We conduct all experiments using 2 NVIDIA 4090 GPUs. For all experiments, we conduct each once using the same random seed. 

\subsection{Detecting Benchmarking Results}
\noindent\textbf{Specialized deep learning models demonstrate superior efficacy but rely on temporal dynamics.} 
As shown in Table \ref{tab:base_results}, among video-level methods, DeMamba \cite{chen2024demamba} leads in overall detection performance (80.69\% ACC, 92.63\% AUC), significantly outperforming image-level baselines. Even the best image-based specialized method, NPR \cite{DBLP:conf/cvpr/TanLZWGLW24}, lags behind with 63.00\% ACC and 87.48\% AUC, indicating that temporal inconsistency remains a primary vulnerability in AIGC videos.
In contrast, LVMs reveal significant gaps in zero-shot detection capabilities. While GPT-4o \cite{islam2024gpt} excels among foundation models (57.25\% ACC, 75.92\% AUC), it still falls short of the specialized TALL detector (77.26\% ACC). Other LVMs show severe limitations, with Claude 3.7 achieving only 23.68\% ACC, failing to distinguish generated content from real-world footage.

\noindent\textbf{Robustness analysis reveals distinct stability profiles under stricter constraints.} 
As shown in Table \ref{tab:accuracy_methods}, most models exhibit performance degradation as the requirement for frame-level consistency increases (from $acc@0.6$ to $acc@1.0$). GPT-4o drops precipitously from 54.25\% to 25.75\% at the strictest threshold ($acc@1.0$), suggesting it relies on sparse artifacts rather than consistent detection across all frames. Gemini-2.5-pro shows similar volatility, degrading sharply from 39.50\% to 14.25\%. However, Qwen-VL-max shows a special strength in stability: despite the lower initial accuracy, it maintains a robust 40.25\% accuracy at the strictest threshold, surpassing GPT-4o (25.75\%).

\subsection{Backtracking Benchmarking Results}
We evaluate the effectiveness of different pre-trained vision backbones on the proposed backtracking task. Table~\ref{tab:backtracking_backbone} reports the retrieval performance using Recall@K (R@1, R@3) and MRR.

\begin{figure*}
  \centering
   \includegraphics[width=0.9\linewidth]{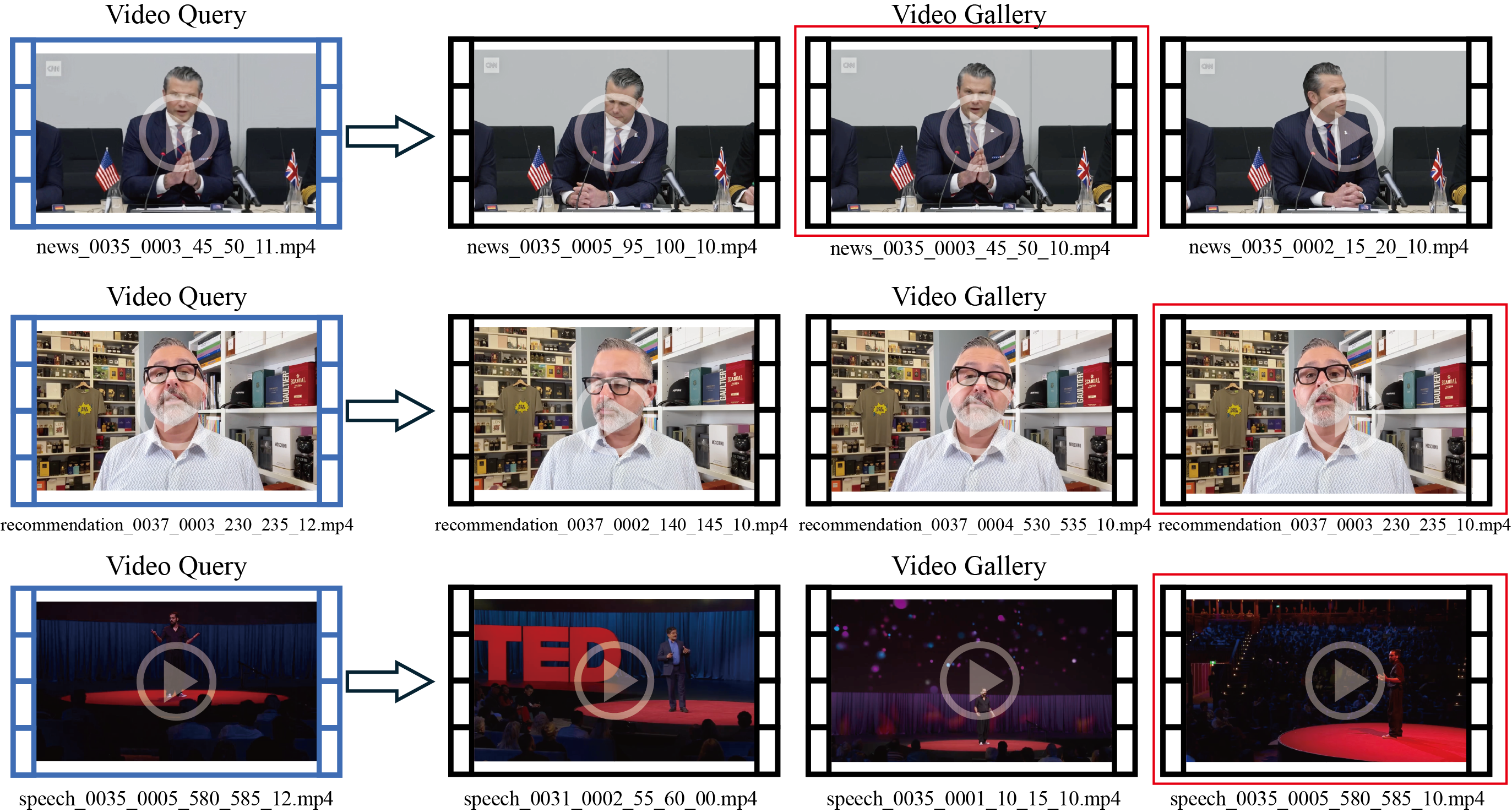}

   \caption{Cases of dino-vit-base-patch16 backtracking task of three types of AI-generated videos in the Chameleon dataset. The red box highlights the candidate retrieved by our model that exactly corresponds to the ground truth.}
   \label{fig:tracking_case}
\end{figure*}

\noindent\textbf{Overall Performance Analysis.}
The quantitative results on the Chameleon dataset demonstrate that identifying the exact real-world source of an AI-generated video is a non-trivial task.
While the best-performing backbone achieves a respectable $Mean_{R@3}$ of 0.900, the $Mean_{R@1}$ drops notably to 0.733.
This significant performance gap reveals a critical characteristic of the backtracking task: while high-level semantic features are generally sufficient to retrieve a shortlist of likely candidates, they often lack the fine-grained discriminative power to pinpoint the definitive source.
Despite the strong performance of these foundation models, the best $Mean_{R@1}$ remains at roughly 73\%. This indicates that identifying the exact source video remains a challenging problem, particularly when the generated query video suffers from temporal misalignment and generation artifacts. The gap between $R@1$ and $R@3$ suggests that models often retrieve the correct source within the top candidates but struggle to pinpoint it as the definitive match.

\noindent\textbf{Theme-specific Analysis.} 
To investigate how content diversity impacts backtracking robustness, we analyze the retrieval performance across three distinct themes: News, Speech, and Recommendation. 
Specifically, News videos achieve remarkable robustness (e.g., $N_{R@1}$ exceeding 80\% for top performers), primarily due to static visual anchors—such as text banners and studio layouts—that survive temporal pruning and provide strong discriminative cues. 
Similarly, Speech content maintains competitive performance ($S_{R@1} \approx 77.5\%$) as unique speaker identities and consistent facial features provide stable semantic consistency across frames. 
In contrast, the Recommendation theme suffers a sharp performance drop, with $R_{R@1}$ often falling below 50\%. 
This degradation highlights a critical task-specific hurdle: unlike the distinct semantics in News or Speech, recommendation videos frequently share generic backgrounds and similar objects. This high inter-class visual ambiguity makes fine-grained source identification extremely difficult, especially when temporal alignment is disrupted.

\subsection{Case Study}
To provide qualitative insights into the model behaviors on the Chameleon dataset, we visualize representative examples for both the detection and backtracking tasks.

\noindent\textbf{Analysis of LVM-based Detection Reasoning.}
Figure~\ref{fig:LVM} illustrates the zero-shot detection capabilities of GPT-4o using our designed prompt. 
As shown in the left column, when the model identifies AI-generated videos, it provides detailed reasoning grounded in specific visual artifacts. 
For instance, in the Text to Video example (middle row), GPT-4o successfully spots the ``distorted fingers merged into one another'', a common failure case in current video generation models. Similarly, in the Digital Human case (bottom row), it detects inconsistent lighting and reflections on the background wall. However, the right column highlights the limitations of image-level LVMs: when the generation quality is high and artifacts are subtle, LVMs tend to fail (predicting ``No''). This visualizes our quantitative finding that while LVMs possess strong semantic understanding, they lack the sensitivity required for robust deepfake detection compared to specialized video-level baselines.

\noindent\textbf{Visualizing the Backtracking Process.}
Figure~\ref{fig:tracking_case} visualizes the retrieval results of the top-performing backbone, DINO-ViT-Base, across three diverse themes: News, Recommendation, and Speech. 
The core challenge of this task is evident in the Gallery: the candidate pool contains multiple ``hard negatives'' real-world videos that feature the same person, scene, or background but differ slightly in timeline or pose. 
For example, in the Recommendation case (middle row), the gallery contains several clips of the same reviewer in the same room. Despite the high semantic overlap between candidates, the model successfully identifies the specific ground-truth video (highlighted in the red box) that corresponds to the generated query. 
This confirms that the learned features capture not just general semantic categories, but also the specific spatiotemporal fingerprints required to establish a one-to-one correspondence between the AI-generated content and its authentic source.

\section{Conclusion}
Current AGVD benchmarks lag behind the rapid evolution of commercial LVMs, failing to capture the high-fidelity visual quality and complex spatiotemporal features of modern synthesized content. To bridge this critical gap, we present the Chameleon dataset and a novel backtracking task. Chameleon uniquely integrates native high-resolution outputs and 3D consistency metrics, enabling a rigorous evaluation of detection methods on realistic, commercial-grade synthetic videos. We have extensively explored detection performance using baselines while simultaneously established a solid methodological and data foundation for the source backtracking of AI-generated videos. In future work, we will focus on extending detection capabilities to longer video sequences and refining the backtracking mechanism to precisely locate original footage, thereby enhancing the forensic community's ability to combat disinformation rooted in advanced deepfakes.

\section*{Acknowledgements}
This work was supported by the Guangdong Engineering Research Center of Data Security Governance and Privacy Computing (No. DSGPC20250101).

\bibliographystyle{ACM-Reference-Format}
\bibliography{sample-base}
\end{document}